\documentclass[10pt,twocolumn,letterpaper]{article}

\usepackage{iccv,times,epsfig,graphicx,amsmath,amssymb,xcolor,tabulary,xfrac,enumitem}
\usepackage[british,english,american]{babel}
\usepackage[pagebackref=true,breaklinks=true,letterpaper=true,colorlinks,
  citecolor=citecolor,bookmarks=false]{hyperref}
\usepackage[caption=false]{subfig}
\definecolor{citecolor}{RGB}{34,139,34}
\def\iccvPaperID{}\iccvfinalcopy

\renewcommand{\paragraph}[1]{\vspace{1.5mm}\noindent\textbf{#1}}
\DeclareMathOperator{\KS}{KS}
\newcommand{\app}{\raise.17ex\hbox{$\scriptstyle\sim$}}
\newcommand{\slname}[1]{\texttt{\small #1}}
\newcommand{\dsname}[1]{\texttt{\small #1}}
\newcommand{\eqnsm}[3]{\vspace{#1}\begin{equation}\label{eq:#2}#3\vspace{#1}\end{equation}\ignorespaces}
\newcommand{\tablestyle}[2]{\setlength{\tabcolsep}{#1}\renewcommand{\arraystretch}{#2}\centering\footnotesize}
\newlength\savewidth\newcommand\shline{\noalign{\global\savewidth\arrayrulewidth
  \global\arrayrulewidth 1pt}\hline\noalign{\global\arrayrulewidth\savewidth}}
\def\x{\times}

\begin{document}
\title{On Network Design Spaces for Visual Recognition\vspace{-3mm}}
\author{%
 Ilija Radosavovic \quad Justin Johnson \quad Saining Xie \quad Wan-Yen Lo \quad Piotr Doll\'ar\\[2mm]
 Facebook AI Research (FAIR)}
\maketitle

\begin{abstract}
Over the past several years progress in designing better neural network architectures for visual recognition has been substantial. To help sustain this rate of progress, in this work we propose to reexamine the methodology for comparing network architectures. In particular, we introduce a new comparison paradigm of \emph{distribution estimates}, in which network design spaces are compared by applying statistical techniques to populations of sampled models, while controlling for confounding factors like network complexity. Compared to current methodologies of comparing point and curve estimates of model families, distribution estimates paint a more complete picture of the entire design landscape. As a case study, we examine design spaces used in neural architecture search (NAS). We find significant statistical differences between recent NAS design space variants that have been largely overlooked. Furthermore, our analysis reveals that the design spaces for standard model families like ResNeXt can be comparable to the more complex ones used in recent NAS work. We hope these insights into distribution analysis will enable more robust progress toward discovering better networks for visual recognition.
\end{abstract}

\section{Introduction}

Our community has made substantial progress toward designing better convolutional neural network architectures for visual recognition tasks over the past several years. This overall research endeavor is analogous to a form of \emph{stochastic gradient descent} where every new proposed model architecture is a noisy gradient step traversing the infinite-dimensional landscape of possible neural network designs. The overall objective of this optimization is to find network architectures that are easy to optimize, make reasonable tradeoffs between speed and accuracy, generalize to many tasks and datasets, and overall withstand the test of time. To make continued steady progress toward this goal, we must use the right \emph{loss function} to guide our search---in other words, a research methodology for comparing network architectures that can reliably tell us whether newly proposed models are truly better than what has come before.

\begin{figure}[t]\centering
\includegraphics[width=1\linewidth]{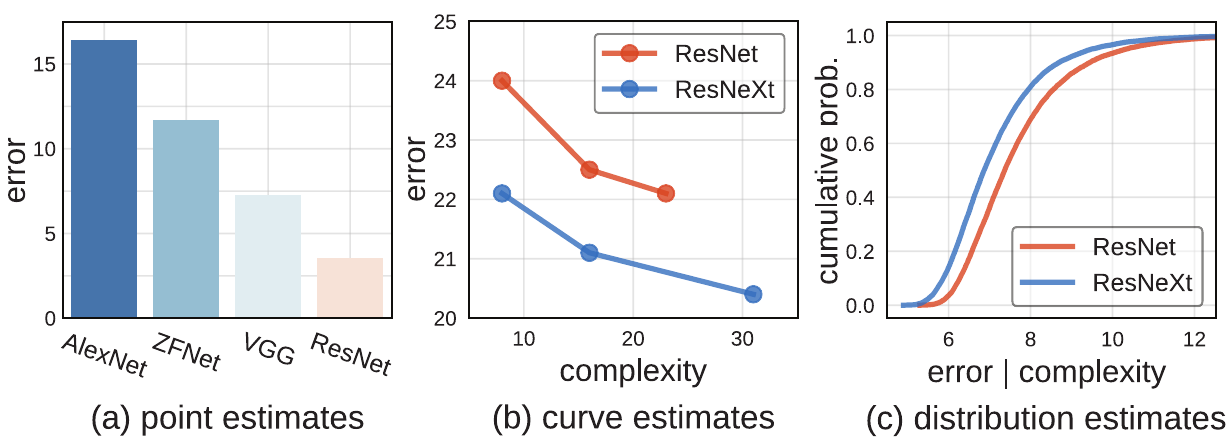}
\caption{\textbf{Comparing networks.} (a) Early work on neural networks for visual recognition tasks used \emph{point estimates} to compare architectures, often \emph{irrespective of model complexity}. (b) More recent work compares \emph{curve estimates} of error \vs complexity traced by a handful of selected models. (c) We propose to \emph{sample} models from a parameterized model design space, and measure \emph{distribution estimates} to compare design spaces. This methodology allows for a more complete and unbiased view of the design landscape.}
\label{fig:teaser}
\end{figure}

One promising avenue lies in developing better \emph{theoretical} understanding of neural networks to guide the development of novel network architectures. However, we need not wait for a general theory of neural networks to emerge to make continued progress. Classical statistics provides tools for drawing informed conclusions from \emph{empirical} studies, even in the absence of a generalized theory governing the subject at hand. We believe that making use of such statistically grounded scientific methodologies in deep learning research may facilitate our future progress.

Overall, there has already been a general trend toward better empiricism in the literature on network architecture design. In the simplest and earliest methodology in this area (Figure~\ref{fig:teaser}a), progress was marked by simple \emph{point estimates}: an architecture was deemed superior if it achieved lower error on a benchmark dataset~\cite{Krizhevsky2012, Lin2013, Zeiler2014, Simonyan2015}, often \emph{irrespective of model complexity}.

An improved methodology adopted in more recent work compares \emph{curve estimates} (Figure~\ref{fig:teaser}b) that explore design tradeoffs of network architectures by instantiating a handful of models from a loosely defined \emph{model family} and tracing curves of error \vs model complexity~\cite{Xie2017, Huang2017dense, Zoph2018}. A model family is then considered superior if it achieves lower error at every point along such a curve. Note, however, that in this methodology other \emph{confounding factors} may vary between model families or may be suboptimal for one of them.

Comparing model families while varying a \emph{single degree of freedom} to generate curve estimates hints at a more general methodology. For a given model family, rather than varying a single network hyperparameter (\eg network depth) while keeping all others fixed (\eg stagewise width, groups), what if instead \emph{we vary all relevant network hyperparameters}? While in principle this would remove confounding factors that may affect conclusions about a model family, it would also yield a vast---often infinite---number of possible models. Are comparisons of model families under such unconstrained conditions even feasible?

To move toward such more robust settings, we introduce a new comparison paradigm: that of \emph{distribution estimates} (Figure~\ref{fig:teaser}c). Rather than comparing a few selected members of a model family, we instead \emph{sample} models from a \emph{design space} parameterizing possible architectures, giving rise to \emph{distributions} of error rates and model complexities. We then compare network design spaces by applying statistical techniques to these distributions, while controlling for confounding factors like network complexity. This paints a more complete and unbiased picture of the design landscape than is possible with point or curve estimates.

To validate our proposed methodology we perform a large-scale empirical study, \emph{training over 100,000 models} spanning multiple model families including VGG~\cite{Simonyan2015}, ResNet~\cite{He2016}, and ResNeXt~\cite{Xie2017} on CIFAR~\cite{Krizhevsky2009}. This large set of trained models allows us to perform \emph{simulations} of distribution estimates and draw robust conclusions about our methodology. In practice, however, we show that sampling between 100 to 1000 models from a given model family is sufficient to perform robust estimates. We further validate our estimates by performing a study on ImageNet~\cite{Deng2009}. This makes the proposed methodology feasible under typical settings and thus a practical tool that can be used to aid in the discovery of novel network architectures.

As a case study of our methodology, we examine the network design spaces used by several recent methods for neural architecture search (NAS)~\cite{Zoph2018, Real2018, Liu2018, Pham2018, Liu2019}. Surprisingly, we find that there are significant differences between the design spaces used by different NAS methods, and we hypothesize that these differences may explain some of the performance improvements between these methods. Furthermore, we demonstrate that design spaces for standard model families such as ResNeXt~\cite{Xie2017} can be comparable to the more complex ones used in recent NAS methods.

We note that our work complements NAS. Whereas NAS is focused on finding the single best model in a given model family, our work focuses on characterizing the model family itself. In other words, our methodology may enable research into \emph{designing the design space} for model search.

We will release the code, baselines, and statistics for all tested models so that proposed future model architectures can compare against the design spaces we consider.

\section{Related Work}

\paragraph{Reproducible research.} There has been an encouraging recent trend toward better reproducibility in machine learning~\cite{Melis2018, Lucic2018, Henderson2018}. For example, Henderson \etal~\cite{Henderson2018} examine recent research in reinforcement learning (RL) and propose guidelines to improve reproducibility and thus enable continued progress in the field. Likewise, we share the goal of introducing a more robust methodology for evaluating model architectures in the domain of visual recognition.

\paragraph{Empirical studies.} In the absence of rigorous theoretical understanding of deep networks, it is imperative to perform large-scale empirical studies of deep networks to aid development~\cite{Greff2015, Collins2017, Novak2018}. For example, in natural language processing, recent large-scale studies~\cite{Melis2018, Merity2018} demonstrate that when design spaces are well explored, the original LSTM~\cite{Hochreiter1997} can outperform more recent models on language modeling benchmarks. These results suggest the crucial role empirical studies and robust methodology play in enabling progress toward discovering better architectures.

\paragraph{Hyperparameter search.} General hyperparameter search techniques~\cite{Bergstra2011, Snoek2012} address the laborious model tuning process in machine learning. A possible approach for comparing networks from two different model families is to first tune their hyperparameters~\cite{Larochelle2007}. However, such comparisons can be challenging in practice. Instead,~\cite{Bergstra2012} advocates using random search as a strong baseline for hyperparameter search and suggests that it additionally helps improve reproducibility. In our work we propose to directly compare the \emph{full} model distributions (not just their minima).

\paragraph{Neural architecture search.} Recently, NAS has proven effective for learning networks architectures~\cite{Zoph2017}. A NAS instantiation has two components: a network \emph{design space} and a \emph{search algorithm} over that space. Most work on NAS focuses on the \emph{search algorithm}, and various search strategies have been studied, including RL~\cite{Zoph2017, Zoph2018, Pham2018}, heuristic search~\cite{Liu2018}, gradient-based search~\cite{Liu2019, Luo2018}, and evolutionary algorithms~\cite{Real2018}. Instead, in our work, we focus on characterizing the model \emph{design space}. As a case study we analyze recent NAS design spaces~\cite{Zoph2018, Real2018, Liu2018, Pham2018, Liu2019} and find significant differences that have been largely overlooked.

\paragraph{Complexity measures.} In this work we focus on analyzing network design spaces while controlling for confounding factors like network complexity. While statistical learning theory~\cite{Schapire1990, Vapnik2013} has introduced theoretical notions of complexity of machine learning models, these are often not predictive of neural network behavior~\cite{Zhang2017, Zhang2019}. Instead, we adopt commonly-used network complexity measures, including the number of model parameters or multiply-add operations~\cite{He2015, Xie2017, Huang2017dense, Zoph2018}. Other measures, \eg wall-clock speed~\cite{Huang2017speed}, can easily be integrated into our paradigm.

\section{Design Spaces}\label{sec:ds}

We begin by describing the core concepts defining a design space in \S\ref{sec:ds:define} and give more details about the actual design spaces used in our experiments in \S\ref{sec:ds:instantiation}.

\begin{table}[t]\centering
\tablestyle{4pt}{1.025}
\begin{tabular}{@{}l|c|c@{}}
 stage & operation & output \\\shline
 stem    & 3$\x$3 conv & 32$\x$32$\x$16\\
 stage 1 & \{block\}$\x$d$_1$ & 32$\x$32$\x$w$_1$\\
 stage 2 & \{block\}$\x$d$_2$ & 16$\x$16$\x$w$_2$\\
 stage 3 & \{block\}$\x$d$_3$ & 8$\x$8$\x$w$_3$\\
 head    & pool + fc & 1$\x$1$\x$10\\   
\end{tabular}\hspace{5mm}
\tablestyle{4pt}{1.05}
\begin{tabular}{@{}l|cc@{}}
                     & R-56 & R-110\\\shline
 flops (B)           & 0.13 & 0.26\\
 params (M)          & 0.86 & 1.73\\
 error~\cite{He2016} & 6.97 & 6.61\\
 error [ours]        & 6.22 & 5.91\\
 \multicolumn{3}{c}{~}\\
\end{tabular}\vspace{1mm}
\caption{\textbf{Design space parameterization}. \emph{(Left)} The general network structure for the standard model families used in our work. Each stage consists of a sequence of $d$ blocks with $w$ output channels (block type varies between design spaces). \emph{(Right)} Statistics of ResNet models~\cite{He2016} for reference. In our notation, R-56 has $d_i{=}9$ and $w_i{=}8{\cdot}2^i$ and R-110 doubles the blocks per stage $d_i$. We report original errors from~\cite{He2016} and those in our reproduction.}
\label{tab:arch}
\end{table}

\subsection{Definitions}\label{sec:ds:define}

\paragraph{I. Model family.} A \emph{model family} is a large (possibly infinite) collection of related neural network architectures, typically sharing some high-level architectural structures or design principles (\eg residual connections). Example model families include standard feedforward networks such as ResNets~\cite{He2016} or the NAS model family from~\cite{Zoph2017, Zoph2018}.

\paragraph{II. Design space.} Performing empirical studies on model families is difficult since they are broadly defined and typically not fully specified. As such we make a distinction between abstract model families, and a \emph{design space} which is a concrete set of architectures that can be instantiated from the model family. A design space consists of two components: a parametrization of a model family such that specifying a set of model hyperparameters fully defines a network instantiation and a set of allowable values for each hyperparameter. For example, a design space for the ResNet model family could include a parameter controlling network depth and a limit on its maximum allowable value.

\paragraph{III. Model distribution.} To perform empirical studies of design spaces, we must instantiate and evaluate a set of network architectures. As a design space can contain an exponential number of candidate models, exhaustive evaluation is not feasible. Therefore, we sample and evaluate a fixed set of models from a design space, giving rise to a \emph{model distribution}, and turn to tools from classical statistics for analysis. Any standard distribution, as well as learned distributions like in NAS, can be integrated into our paradigm.

\paragraph{IV. Data generation.} To analyze network design spaces, we sample and evaluate numerous models from each design space. In doing so, we effectively generate \emph{datasets} of trained models upon which we perform empirical studies.

\begin{table}[t]\centering
\tablestyle{5.5pt}{1.05}
\begin{tabular}{@{}l|cccc|c@{}}
 & depth & width & ratio & groups & total\\\shline
 \dsname{Vanilla}   & 1,24,9 & 16,256,12  &       &         & 1,259,712\\
 \dsname{ResNet}    & 1,24,9 & 16,256,12  &       &         & 1,259,712\\
 \dsname{ResNeXt-A} & 1,16,5 & 16,256,5~~ & 1,4,3 & 1,4,3~~ & 11,390,625\\
 \dsname{ResNeXt-B} & 1,16,5 & 64,1024,5  & 1,4,3 & 1,16,5  & 52,734,375\\
\end{tabular}\vspace{1mm}
\caption{\textbf{Design spaces.} Independently for each of the three network stages $i$, we select the number of blocks $d_i$ and the number of channels per block $w_i$. Notation $a$, $b$, $n$ means we sample from $n$ values spaced about evenly (in log space) in the range $a$ to $b$. For the \dsname{ResNeXt} design spaces, we also select the bottleneck width ratio $r_i$ and the number of groups $g_i$ per stage. The total number of models is $(dw)^3$ and $(dwrg)^3$ for models w/o and with groups.}
\label{tab:design_spaces}\vspace{-2mm}
\end{table}

\subsection{Instantiations}\label{sec:ds:instantiation}

We provide precise description of design spaces used in the analysis of our methodology. We introduce additional design spaces for NAS model families in \S\ref{sec:nas}.

\paragraph{I. Model family.} We study three standard model families. We consider a `vanilla' model family (feedforward network loosely inspired by VGG~\cite{Simonyan2015}), as well as model families based on ResNet~\cite{He2016} and ResNeXt~\cite{Xie2017} which use residuals and grouped convolutions. We provide more details next.

\paragraph{II. Design space.} Following~\cite{He2016}, we use networks consisting of a stem, followed by three stages, and a head, see Table~\ref{tab:arch} (left). Each stage consists of a sequence of blocks. For our \dsname{ResNet} design space, a single block consists of two convolutions\footnote{All convs are 3$\x$3 and are followed by Batch Norm~\cite{Ioffe2015} and ReLU.} and a residual connection. Our \dsname{Vanilla} design space uses an identical block structure but without residuals. Finally, in case of the ResNeXt design spaces, we use bottleneck blocks with groups~\cite{Xie2017}. Table~\ref{tab:arch} (right) shows some baseline ResNet models for reference (for details of the training setup see appendix). To complete the design space definitions, in Table~\ref{tab:design_spaces} we specify the set of allowable hyperparameters for each. Note that we consider two variants of the ResNeXt design spaces with different hyperparameter sets: \dsname{ResNeXt-A} and \dsname{ResNeXt-B}.

\paragraph{III. Model distribution.} We generate model distributions by \emph{uniformly} sampling hyperparameter from the allowable values for each design spaces (as specified in Table~\ref{tab:design_spaces}).

\paragraph{IV. Data generation.} Our main experiments use image classification models trained on CIFAR-10~\cite{Krizhevsky2009}. This setting enables large-scale analysis and is often used as a testbed for recognition networks, including for NAS. While we find that sparsely sampling models from a given design space is sufficient to obtain robust estimates, we perform much denser sampling \emph{to evaluate our methodology}. We sample and train 25k models from each of the design spaces from Table~\ref{tab:design_spaces}, for a total of 100k models. To reduce computational load, we consider models for which the flops\footnote{Following common practice, we use flops to mean multiply-adds.} or parameters are below the ResNet-56 values (Table~\ref{tab:arch}, right).

\section{Proposed Methodology}\label{sec:methodology}

In this section we introduce and evaluate our methodology for comparing design spaces. Throughout this section we use the design spaces introduced in \S\ref{sec:ds:instantiation}.

\subsection{Comparing Distributions}\label{sec:compare_distributions}

When developing a new network architecture, human experts employ a combination of grid and manual search to evaluate models from a design space, and select the model achieving the lowest error (\eg as described in~\cite{Larochelle2007}). The final model is a \emph{point estimate} of the design space. As a community we commonly use such point estimates to draw conclusions about which methods are superior to others.

Unfortunately, comparing design spaces via point estimates can be misleading. We illustrate this using a simple example: we consider comparing two sets of models of \emph{different sizes} sampled from the \emph{same design space}.

\paragraph{Point estimates.} As a proxy for human derived point estimates, we use random search~\cite{Bergstra2012}. We generate a \emph{baseline} model set (\slname{B}) by uniformly sampling 100 architectures from our \dsname{ResNet} design space (see Table~\ref{tab:design_spaces}). To generate the second model set (\slname{M}), we instead use 1000 samples. In practice, the difference in number of samples could arise from more effort in the development of \slname{M} over the baseline, or simply access to more computational resources for generating \slname{M}. Such \emph{imbalanced comparisons} are common in practice.

\begin{figure}[t]\centering
\includegraphics[width=1.0\linewidth]{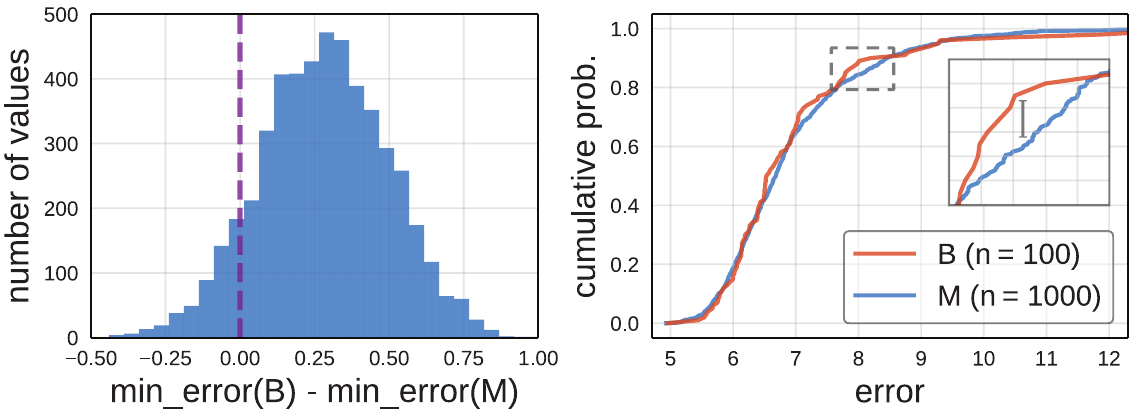}
\caption{\textbf{Point \vs distribution comparison}. Consider two sets of models, \slname{B} and \slname{M}, generated by randomly sampling 100 and 1000 models, respectively, from the \emph{same} design space. Such situations commonly arise in practice, \eg due to more effort being devoted to model development of a new method. \emph{(Left)} The difference in the \emph{minimum} error of \slname{B} and \slname{M} over 5000 random trials. In 90\% of cases \slname{M} has lower minimum than \slname{B}, leading to incorrect conclusions. \emph{(Right)} Comparing EDFs of the errors (Eqn.~\ref{eq:edf}) from \slname{B} and \slname{M} immediately suggests the distributions of the two sets are likely the same. We can quantitatively characterize this by measuring the $\KS$ statistic $D$ (Eqn.~\ref{eq:ks}), computed as the maximum vertical discrepancy between two EDFs, as shown in the zoomed in panel.}
\label{fig:small_large}
\end{figure}

After training, \slname{M}'s \emph{minimum error} is lower than \slname{B}'s \emph{minimum error}. Since the best error is lower, a naive comparison of \emph{point estimates} concludes that \slname{M} is superior. Repeating this experiment yields the same result: Figure~\ref{fig:small_large} (left) plots the difference in the minimum error of \slname{B} and \slname{M} over multiple trials (simulated by repeatedly sampling \slname{B} and \slname{M} from our pool of 25k pre-trained models). In 90\% of cases \slname{M} has a lower minimum than \slname{B}, often by a large margin. Yet clearly \slname{B} and \slname{M} were drawn from the \emph{same} design space, so this analysis based on point estimates can be misleading.

\paragraph{Distributions.} In this work we make the case that one can draw more robust conclusions by directly comparing \emph{distributions} rather than \emph{point estimates} such as minimum error.

To compare distributions, we use \emph{empirical distribution functions} (EDFs). Let $\mathbf{1}$ be the indicator function. Given a set of $n$ models with errors $\{e_i\}$, the error EDF is given by:
 \eqnsm{-1mm}{edf}{F(e)=\frac1n\sum_{i=1}^n \mathbf{1}[e_i < e].}
$F(e)$ gives the fraction of models with error less than $e$.

We revisit our \slname{B} \vs \slname{M} comparison in Figure~\ref{fig:small_large} (right), but this time plotting the full error distributions instead of just their minimums. Their shape is typical of error EDFs: the small tail to the bottom left indicates a small population of models with low error and the long tail on the upper right shows there are few models with error over 10\%.

\emph{Qualitatively}, there is little visible difference between the error EDFs for \slname{B} and \slname{M}, suggesting that these two sets of models were drawn from an identical design space. We can make this comparison \emph{quantitative} using the (two sample) Kolmogorov-Smirnov ($\KS$) test~\cite{Massey1951}, a nonparametric statistical test for the null hypothesis that two samples were drawn from the same distribution. Given EDFs $F_1$ and $F_2$, the test computes the $\KS$ statistic $D$, defined as:
 \eqnsm{0mm}{ks}{D = \sup_x|F_1(x)-F_2(x)|}
$D$ measures the maximum vertical discrepancy between EDFs (see the zoomed in panel in Figure~\ref{fig:small_large}); small values suggest that $F_1$ and $F_2$ are sampled from the same distribution. For our example, the $\KS$ test gives $D=0.079$ and a $p$-value of 0.60, so with high confidence we fail to reject the null hypothesis that \slname{B} and \slname{M} follow the same distribution.

\paragraph{Discussion.} While pedagogical, the above example demonstrates the necessity of comparing distributions rather than point estimates, as the latter can give misleading results in even simple cases. We emphasize that such imbalanced comparisons occur frequently in practice. Typically, most work reports results for only a small number of best models, and rarely reports the number of total points explored during model development, which can vary substantially.

\begin{figure}[t]\centering
\includegraphics[width=1.0\linewidth]{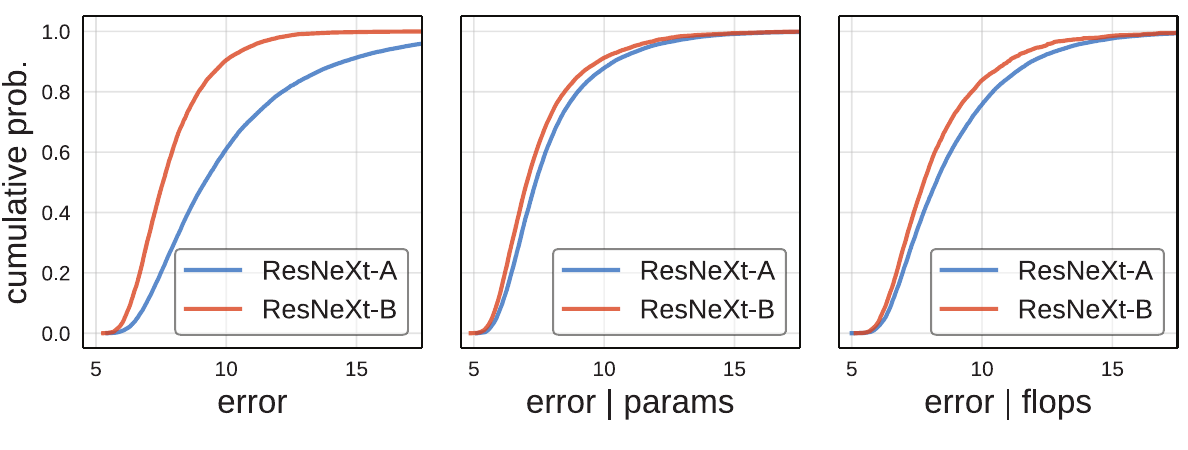}\vspace{-1mm}
\caption{\textbf{Comparisons conditioned on complexity.} Without considering complexity, there is a large gap between the related \dsname{ResNeXt-A} and \dsname{ResNeXt-B} design spaces (left). We control for params (middle) and flops (right) to obtain distributions \emph{conditioned} on complexity (Eqn.~\ref{eq:nedf}), which we denote by `error~$|$~complexity'. Doing so removes most of the observed gap and isolates differences that cannot be explained by model complexity alone.}
\label{fig:complexity_edfs}\vspace{-4mm}
\end{figure}

\subsection{Controlling for Complexity}\label{sec:methodology:complexity}

While comparing distributions can lead to more robust conclusions about design spaces, when performing such comparisons, we need to control for \emph{confounding factors} that correlate with model error to avoid biased conclusions. A particularly relevant confounding factor is \emph{model complexity}. We study controlling for complexity next.

\paragraph{Unnormalized comparison.} Figure~\ref{fig:complexity_edfs} (left) shows the error EDFs for the \dsname{ResNeXt-A} and \dsname{ResNeXt-B} design spaces, which differ only in the allowable hyperparameter sets for sampling models (see Table~\ref{tab:design_spaces}). Nevertheless, the curves have clear qualitative differences and suggest that \dsname{ResNeXt-B} is a better design space. In particular, the EDF for \dsname{ResNeXt-B} is higher; \ie, it has a higher fraction of better models at every error threshold. This clear difference illustrates that different design spaces from the \emph{same model family} under the same model distribution can result in very different error distributions. We investigate this gap further.

\begin{figure}[t]\centering
\includegraphics[width=1.0\linewidth]{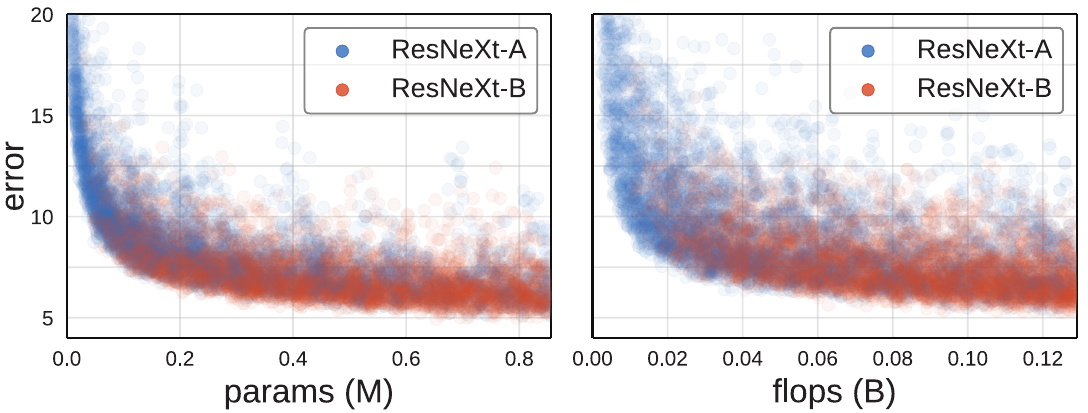}\vspace{-1mm}
\caption{\textbf{Complexity \vs error.} We show the relationship between model complexity and performance for two different complexity metrics and design spaces. Each point is a trained model.}
\label{fig:complexity_vs_error}
\end{figure}

\paragraph{Error vs complexity.} From prior work we know that a model's error is related to its complexity; in particular more complex models are often more accurate~\cite{He2015, Xie2017, Huang2017dense, Zoph2018}. We explore this relationship using our large-scale data. Figure~\ref{fig:complexity_vs_error} plots the error of each trained model against its complexity, measured by either its parameter or flop counts. While there are poorly-performing high-complexity models, the \emph{best models have the highest complexity}. Moreover, in this setting, we see no evidence of saturation: as complexity increases we continue to find better models.

\begin{figure}[t]\centering
\includegraphics[width=1.0\linewidth]{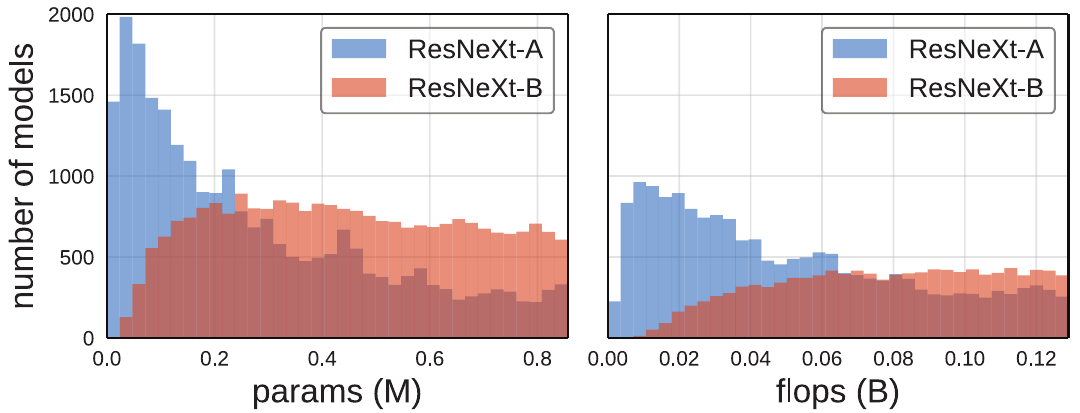}
\caption{\textbf{Complexity distribution.} Two different ResNeXt design spaces have different complexity distributions. We need to account for this difference in order to be able to compare the two.}
\label{fig:complexity_dist}\vspace{-3mm}
\end{figure}

\paragraph{Complexity distribution.} Can the differences between the \dsname{ResNeXt-A} and \dsname{ResNeXt-B} EDFs in Figure~\ref{fig:complexity_edfs} (left) be due to differences in their complexity distributions? In Figure~\ref{fig:complexity_dist}, we plot the empirical distributions of model complexity for the two design spaces. We see that \dsname{ResNeXt-A} contains a much larger number of low-complexity models, while \dsname{ResNeXt-B} contains a heavy tail of high-complexity models. It therefore seems plausible that \dsname{ResNeXt-B}'s apparent superiority is due to the confounding effect of complexity.

\paragraph{Normalized comparison.} We propose a \emph{normalization} procedure to factor out the \emph{confounding} effect of the differences in the complexity of model distributions. Given a set of $n$ models where each model has complexity $c_i$, the idea is to assign to each model a weight $w_i$, where $\Sigma_i w_i{=}1$, to create a more \emph{representative} set under that complexity measure.

Specifically, given a set of $n$ models with errors, complexities, and weights given by $\{e_i\}$, $\{c_i\}$, and $\{w_i\}$, respectively, we define the \emph{normalized complexity} EDF as:
 \eqnsm{-2mm}{cedf}{C(c) = \sum_{i=1}^n w_i \mathbf{1}[c_i < c] }
Likewise, we define the \emph{normalized error} EDF as:
 \eqnsm{-2mm}{nedf}{\hat{F}(e)=\sum_{i=1}^n w_i\mathbf{1}[e_i < e].}
Then, given two model sets, our goal is to find weights for each model set such that $C_1(c){\approx}C_2(c)$ for all $c$ in a given complexity range. Once we have such weights, comparisons between $\hat{F}_1$ and $\hat{F}_2$ reveal differences between design spaces that cannot be explained by model complexity alone.

In practice, we set the weights for a model set such that its complexity distribution (Eqn.~\ref{eq:cedf}) is \emph{uniform}. Specifically, we bin the complexity range into $k$ bins, and assign each of the $m_j$ models that fall into a bin $j$ a weight $w_j=\sfrac{1}{km_j}$. Given sparse data, the assignment into bins could be made in a soft manner to obtain smoother estimates. While other options for matching $C_1{\approx}C_2$ are possible, we found normalizing both $C_1$ and $C_2$ to be uniform to be effective.

In Figure~\ref{fig:complexity_edfs} we show \dsname{ResNeXt-A} and \dsname{ResNeXt-B} error EDFs, normalized by parameters (middle) and flops (right). Controlling for complexity brings the curves closer, suggesting that much of the original gap was due to mismatched complexity distributions. This is not unexpected as the design spaces are similar and both parameterize the same underlying model family. We observe, however, that their normalized EDFs still show a small gap. We note that \dsname{ResNeXt-B} contains wider models with more groups (see Table~\ref{tab:design_spaces}), which may account for this remaining difference.

\subsection{Characterizing Distributions}\label{sec:methodology:characterizing}

\begin{figure}[t]\centering
\includegraphics[width=1.0\linewidth]{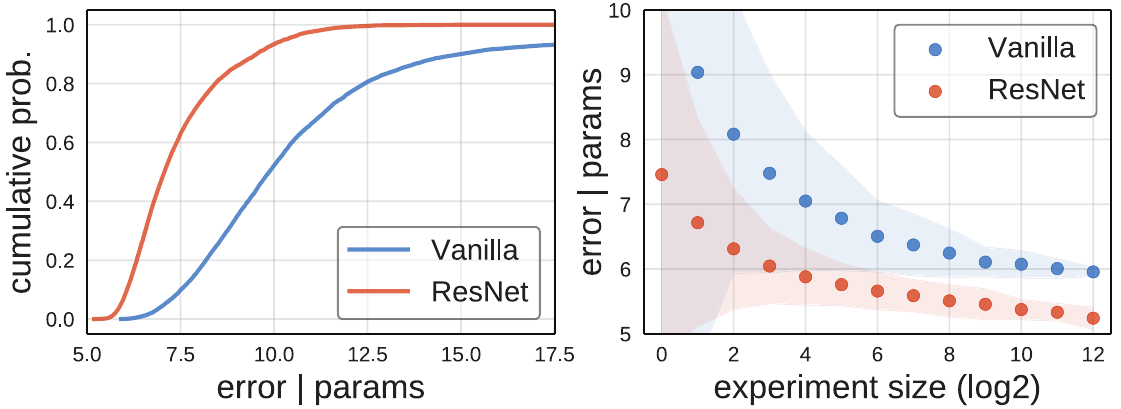}
\caption{\textbf{Finding good models quickly.} \emph{(Left)} The \dsname{ResNet} design space contains a much larger proportion of good models than the \dsname{Vanilla} design space, making \emph{finding a good model quickly} easier. \emph{(Right)} This can also be seen by measuring the number of model evaluations random search requires to reach a given error.}
\label{fig:good_models}\vspace{-3mm}
\end{figure}

An advantage of examining the full error distribution of a design space is it gives insights beyond the minimum achievable error. Often, we indeed focus on finding the \emph{best model} under some complexity constraint, for example, if a model will be deployed in a production system. In other cases, however, we may be interested in \emph{finding a good model quickly}, \eg when experimenting in a new domain or under constrained computation. Examining distributions allows us to more fully characterize a design space.

\paragraph{Distribution shape.} Figure~\ref{fig:good_models} (left) shows EDFs for the \dsname{Vanilla} and \dsname{ResNet} design spaces (see Table~\ref{tab:design_spaces}). In the case of \dsname{ResNet}, the majority ($>$80\%) of models have error under 8\%. In contrast, the \dsname{Vanilla} design space has a much smaller fraction of such models (\app15\%). This makes it easier to find a good \dsname{ResNet} model. While this is not surprising given the well-known effectiveness of residual connections, it does demonstrate how the shape of the EDFs can give additional insight into characterizing a design space.

\paragraph{Distribution area.} We can summarize an EDF by the average area under the curve up to some max $\epsilon$. That is we can compute $\int_0^{\epsilon} \hat{F}(e)/\epsilon~\mathrm{d}e{=}1{-}\sum w_i \min(1,\frac{e_i}{\epsilon})$. For our example, \dsname{ResNet} has a larger area under the curve. However, like the min, the area gives only a partial view of the EDF.


\paragraph{Random search efficiency.} Another way to assess the ease of finding a good model is to measure random search efficiency. To \emph{simulate random search} experiments of varying size $m$ we follow the procedure described in~\cite{Bergstra2012}. Specifically, for each experiment size $m$, we sample $m$ models from our pool of $n$ models and take their \emph{minimum} error. We repeat this process $\app\sfrac{n}{m}$ times to obtain the mean along with error bars for each $m$. To factor out the confounding effect of complexity, we assign a weight to each model such that $C_1{\approx}C_2$ (Eqn.~\ref{eq:cedf}) and use these weights for sampling.

In Figure~\ref{fig:good_models} (right), we use our 50k pre-trained models from the \dsname{Vanilla} and \dsname{ResNet} design spaces to simulate random search (conditioned on parameters) for varying $m$. We observe consistent findings as before: random search finds better models faster in the \dsname{ResNet} design space.

\subsection{Minimal Sample Size}\label{sec:methodology:minimal}

\begin{figure}[t]\centering
\includegraphics[width=1.0\linewidth]{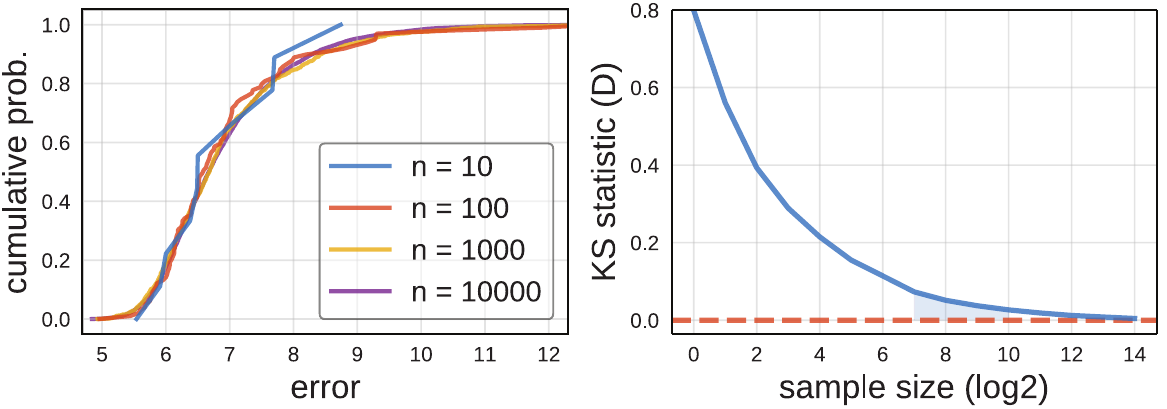}
\caption{\textbf{Number of samples.} We analyze the number of samples required for performing analysis of design spaces using our methodology. \emph{(Left)} We show a qualitative comparison of EDFs generated using a varying number of samples. \emph{(Right)} We compute $\KS$ statistic between the full sample and sub-samples of increasing size. In both cases, we arrive at the conclusion that between 100 and 1000 samples is a reasonable range for our methodology.}
\label{fig:num_samples}\vspace{-2mm}
\end{figure}

Our experiments thus far used very large sets of trained models. In practice, however, far fewer samples can be used to compare distributions of models as we now demonstrate.

\paragraph{Qualitative analysis.} Figure~\ref{fig:num_samples} (left) shows EDFs for the \dsname{ResNet} design space with varying number of samples. Using 10 samples to generate the EDF is quite noisy; however, 100 gives a reasonable approximation and 1000 is visually indistinguishable from 10,000. This suggests that 100 to 1000 samples may be sufficient to compare distributions.

\paragraph{Quantitative analysis.} We perform quantitative analysis to give a more precise characterization of the number of samples necessary to compare distributions. In particular, we compute the $\KS$ statistic $D$ (Eqn.~\ref{eq:ks}) between the full sample of 25k models and sub-samples of increasing size $n$. The results are shown in Figure~\ref{fig:num_samples} (right). As expected, as $n$ increases, $D$ decreases. At 100 samples $D$ is about 0.1, and at 1000 $D$ begins to saturate. Beyond 1000 samples shows diminishing returns. Thus, our earlier estimate of 100 samples is indeed a reasonable lower-bound, and 1000 should be sufficient for more precise comparisons. We note, however, that these bounds may vary under other circumstances.

\paragraph{Feasibility discussion.} One might wonder about the feasibility of training between 100 and 1000 models for evaluating a distribution. In our setting, training 500 CIFAR models requires about 250 GPU hours. In comparison, training a typical ResNet-50 baseline on ImageNet requires about 192 GPU hours. Thus, evaluating the full distribution for a small-sized problem like CIFAR requires a computational budget on par with a point estimate for a medium-sized problem like ImageNet. To put this in further perspective, NAS methods can require as much as $O(10^5)$ GPU hours on CIFAR~\cite{Pham2018}. Overall, we expect distribution comparisons to be quite feasible under typical settings. To further aid such comparisons, we will release data for all studied design spaces to serve as baselines for future work.

\begin{table}[t]
\tablestyle{8pt}{1.05}
\begin{tabular}{l|ccc|r}
 & \#ops & \#nodes & output & \#cells (B)\\\shline
 NASNet~\cite{Zoph2018} & 13 & 5 & L & 71,465,842\\
 Amoeba~\cite{Real2018} &  8 & 5 & L & 556,628\\
 PNAS~\cite{Liu2018}    &  8 & 5 & A & 556,628\\
 ENAS~\cite{Pham2018}   &  5 & 5 & L & 5,063\\
 DARTS~\cite{Liu2019}   &  8 & 4 & A & 242\\
\end{tabular}\vspace{1mm}
\caption{\textbf{NAS design spaces.} We summarize the \emph{cell structure} for five NAS design spaces. We list the number of candidate ops (\eg 5$\x$5 conv, 3$\x$3 max pool, \etc), number of nodes (excluding the inputs), and which nodes are concatenated for the output (`A' if `all' nodes, `L' if `loose' nodes not used as input to other nodes). Given $o$ ops to choose from, there are $o^2{\cdot}(j{+}1)^2$ choices when adding the $j^{th}$ node, leading to $o^{2k}{\cdot}((k{+}1)!)^2$ possible cells with $k$ nodes (of course many of these cells are redundant). The spaces vary substantially; indeed, even exact candidate ops for each vary.}
\label{tab:nas_details}\vspace{-4mm}
\end{table}

\section{Case Study: NAS}\label{sec:nas}

As a case study of our methodology we examine design spaces from recent neural architecture search (NAS) literature. In this section we perform studies on CIFAR~\cite{Krizhevsky2009} and in Appendix B we further validate our results by replicating the study on ImageNet~\cite{Deng2009}, yielding similar conclusions.

NAS has two core components: a \emph{design space} and a \emph{search algorithm} over that space. While normally the focus is on the search algorithm (which can be viewed as inducing a distribution over the design space), we instead focus on comparing the design spaces under a fixed distribution. Our main finding is that in recent NAS papers, significant design space differences have been largely overlooked. Our approach complements NAS by \emph{decoupling} the design of the design space from the design of the search algorithm, which we hope will aid the study of new design spaces.

\subsection{Design Spaces}\label{sec:nas:ds}

\paragraph{I. Model family.} The general NAS model family was introduced in~\cite{Zoph2017, Zoph2018}. A NAS model is constructed by repeatedly stacking a single computational unit, called a \emph{cell}, where a cell can vary in the operations it performs and in its connectivity pattern. In particular, a cell takes outputs from two previous cells as inputs and contains a number of \emph{nodes}. Each node in a cell takes as input two previously constructed nodes (or the two cell inputs), applies an operator to each input (\eg convolution), and combines the output of the two operators (\eg by summing). We refer to ENAS~\cite{Pham2018} for a more detailed description.

\paragraph{II. Design space.} In spite of many recent papers using the general NAS model family, most recent approaches use different design space instantiations. In particular, we carefully examined the design spaces described in NASNet~\cite{Zoph2018}, AmoebaNet~\cite{Real2018}, PNAS~\cite{Liu2018}, ENAS~\cite{Pham2018}, and DARTS~\cite{Liu2019}. The \emph{cell structure} differs substantially between them, see Table~\ref{tab:nas_details} for details. In our work, we define five design spaces by reproducing these five cell structures, and name them accordingly, \ie, \dsname{NASNet}, \dsname{Amoeba}, \etc.

\begin{figure}[t]\centering
\includegraphics[width=1.00\linewidth]{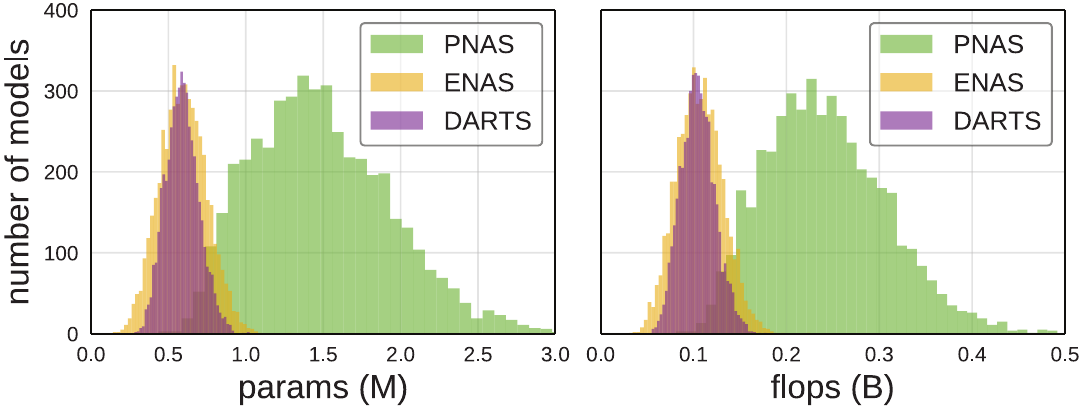}
\caption{\textbf{NAS complexity distribution.} Complexity of design spaces with different cell structures (see Table~\ref{tab:nas_details}) vary significantly given fixed width ($w{=}$16) and depth ($d{=}$20). To compare design spaces we need to normalize for complexity, which requires the complexity distributions to fall in the same range. We achieve this by allowing $w$ and $d$ to vary (and setting a max complexity).}
\label{fig:nas_complexity_dist}\vspace{-2mm}
\end{figure}

How the cells are stacked to generate the full \emph{network architecture} also varies slightly between recent papers, but less so than the cell structure. We therefore standardize this aspect of the design spaces; that is we adopt the network architecture setting from DARTS~\cite{Liu2019}. Core aspects include the stem structure, even placement of the three reduction cells, and filter width that doubles after each reduction cell.

The network depth $d$ and initial filter width $w$ are typically kept fixed. However, these hyperparameters directly affect model complexity. Specifically, Figure~\ref{fig:nas_complexity_dist} shows the complexity distribution generated with different cell structures with $w$ and $d$ kept fixed. The ranges of the distributions differ due to the varying cell structures designs. To factor out this confounding factor, we let $w$ and $d$ vary (selecting $w\in\{16,24,32\}$ and $d\in\{4,8,12,16,20\}$). This spreads the range of the complexity distributions for each design space, allowing for more controlled comparisons.

\paragraph{III. Model distribution.} We sample NAS cells by using uniform sampling at each step (\eg operator and node selection). Likewise, we sample $w$ and $d$ uniformly at random.

\paragraph{IV. Data generation.} We train $\app$1k models on CIFAR for each of the five NAS design spaces in Table~\ref{tab:nas_details} (see \S\ref{sec:methodology:minimal} for a discussion of sample size). In particular, we ensure we have 1k models per design space for both the full flop range and the full parameter range (upper-bounded by R-56, see \S\ref{sec:ds}).

\subsection{Design Space Comparisons}\label{sec:nas:experiments}

We adopt our distribution comparison tools (EDFs, $\KS$ test, \etc) from \S\ref{sec:methodology} to compare the five NAS design spaces, each of which varies in its cell structure (see Table~\ref{tab:nas_details}).

\begin{figure}[t]\centering
\includegraphics[width=1.0\linewidth]{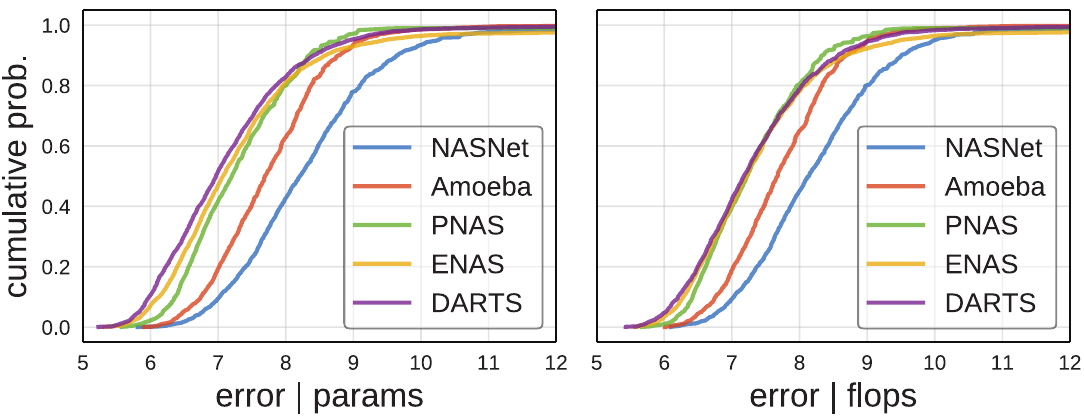}
\caption{\textbf{NAS distribution comparisons.} EDFs for five NAS design spaces from Table~\ref{tab:nas_details}. The EDFs differ substantially (max $\KS$ test $D=0.51$ between \dsname{DARTS} and \dsname{NASNet}) even though the design spaces are all instantiations of the NAS model family.}
\label{fig:nas_edfs}\vspace{-1mm}
\end{figure}

\paragraph{Distribution comparisons.} Figure~\ref{fig:nas_edfs} shows normalized error EDFs for each of the NAS design spaces. Our main observation is that the EDFs vary considerably: the \dsname{NASNet} and \dsname{Amoeba} design spaces are noticeably worse than the others, while \dsname{DARTS} is best overall. Comparing \dsname{ENAS} and \dsname{PNAS} shows that while the two are similar, \dsname{PNAS} has more models with intermediate errors while \dsname{ENAS} has more lower/higher performing models, causing the EDFs to cross.

Interestingly, according to our analysis the design spaces corresponding to newer work outperform the earliest design spaces introduced in NASNet~\cite{Zoph2018} and Amoeba~\cite{Real2018}. While the NAS literature typically focuses on the search algorithm, the design spaces also seem to be improving. For example, PNAS~\cite{Liu2018} removed five ops from NASNet that were not selected in the NASNet search, effectively pruning the design space. Hence, at least part of the gains in each paper may come from improvements of the design space.

\begin{figure}[t]\centering
\includegraphics[width=1.0\linewidth]{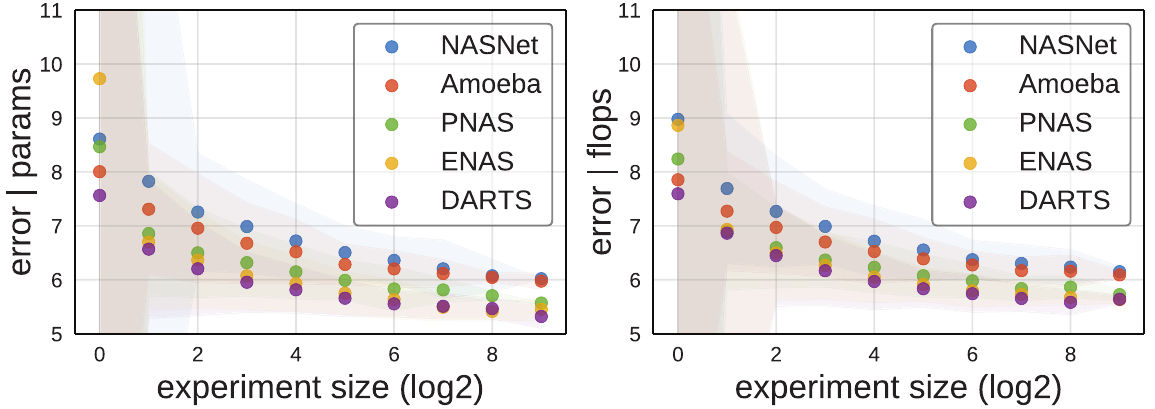}
\caption{\textbf{NAS random search efficiency.} Design space differences lead to clear differences in the efficiency of random search on the five tested NAS design spaces. This highlights the importance of decoupling the \emph{search algorithm} and \emph{design space}.}
\label{fig:nas_rs}\vspace{-3mm}
\end{figure}

\paragraph{Random search efficiency.} We simulate random search in the NAS design spaces (after normalizing for complexity) following the setup from \S\ref{sec:methodology:characterizing}. Results are shown in Figure~\ref{fig:nas_rs}. First, we observe that ordering of design spaces by random search efficiency is consistent with the ordering of the EDFs in Figure~\ref{fig:nas_edfs}. Second, for a fixed search algorithm (random search in this case), this shows the differences in the design spaces alone leads to clear differences in performance. This reinforces that care should be taken to keep the design space fixed if the search algorithm is varied.

\subsection{Comparisons to Standard Design Spaces}

\begin{figure}[t]\centering
\includegraphics[width=1.0\linewidth]{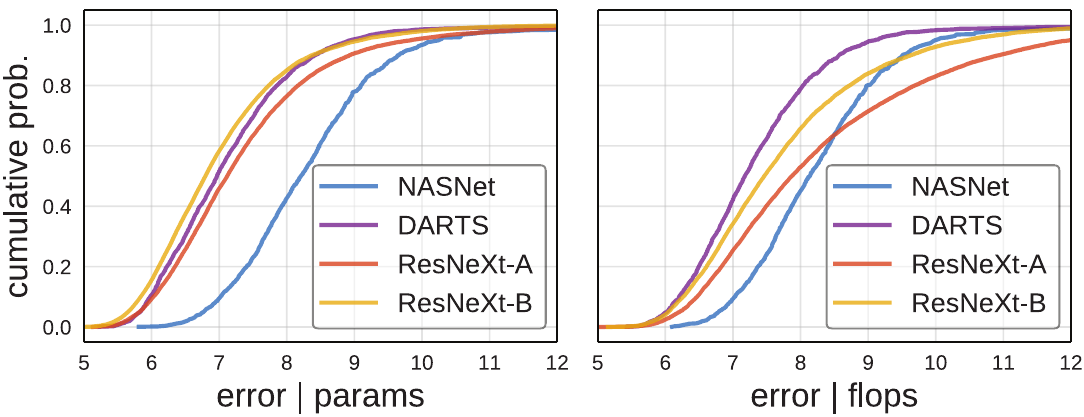}
\caption{\textbf{NAS \vs standard design spaces.} Normalized for params, \dsname{ResNeXt-B} is comparable to the strong \dsname{DARTS} design space ($\KS$ test $D{=}0.09$). Normalized for flops, \dsname{DARTS} outperforms  \dsname{ResNeXt-B}, but not by a relatively large margin.}
\label{fig:nas_std_edfs}\vspace{-1mm}
\end{figure}

We next compare the NAS design spaces with the design spaces from \S\ref{sec:ds}. We select the best and worst performing NAS design spaces (\dsname{DARTS} and \dsname{NASNet}) and compare them to the two ResNeXt design spaces from Table~\ref{tab:design_spaces}. EDFs are shown in Figure~\ref{fig:nas_std_edfs}. \dsname{ResNeXt-B} is on par with \dsname{DARTS} when normalizing by params (left), while \dsname{DARTS} slightly outperforms \dsname{ResNeXt-B} when normalizing by flops (right). \dsname{ResNeXt-A} is worse than \dsname{DARTS} in both cases.

It is interesting that ResNeXt design spaces \emph{can be} comparable to the NAS design spaces (which vary in cell structure in addition to width and depth). These results demonstrate that the \emph{design} of the design space plays a key role and suggest that \emph{designing design spaces}, manually or via data-driven approaches, is a promising avenue for future work.

\begin{table}[t]
\tablestyle{6pt}{1.05}
\begin{tabular}{l|cc|ccc}
 & flops & params & error & error & error\\[-.5mm]
 &  (B) & (M) & original & default & enhanced \\\shline
 ResNet-110      & 0.26 & 1.7 & 6.61 & 5.91 & 3.65\\
 ResNeXt$^\star$ & 0.38 & 2.5 & --   & 4.90 & 2.75\\
 DARTS$^\star$   & 0.54 & 3.4 & 2.83 & 5.21 & 2.63\\
\end{tabular}\vspace{1mm}
\caption{\textbf{Point comparisons.} We compare selected higher complexity models using \emph{originally} reported errors to our \emph{default} training setup and the \emph{enhanced} training setup from~\cite{Liu2019}. The results show the importance of carefully controlling the training setup.}
\label{tab:best_models}\vspace{-3mm}
\end{table}

\subsection{Sanity Check: Point Comparisons}

We note that recent NAS papers report lower overall errors due to higher complexity models and \emph{enhanced} training settings. As a \emph{sanity check}, we perform \emph{point comparisons} using larger models and the \emph{exact} training settings from DARTS~\cite{Liu2019} which uses a 600 epoch schedule with deep supervision~\cite{Lee2015}, Cutout~\cite{Devries2017}, and modified DropPath~\cite{Larsson2017}. We consider three models: DARTS$^\star$ (the best model found in DARTS~\cite{Liu2019}), ResNeXt$^\star$ (the best model from \dsname{ResNeXt-B} with increased widths), and ResNet-110~\cite{He2016}.

Results are shown in Table~\ref{tab:best_models}. With the \emph{enhanced} setup, ResNeXt$^\star$ achieves similar error as DARTS$^\star$ (with comparable complexity). This reinforces that performing comparisons under the same settings is crucial, simply using an enhanced training setup gives over 2\% gain; even the original ResNet-110 is competitive under these settings.

\section{Conclusion}

We present a methodology for analyzing and comparing model design spaces. Although we focus on convolutional networks for image classification, our methodology should be applicable to other model types (\eg RNNs), domains (\eg NLP), and tasks (\eg detection). We hope our work will encourage the community to consider design spaces as a core part of model development and evaluation.

\section*{Appendix A: Supporting Experiments}

In the appendix we provide details about training settings and report extra experiments to verify our methodology.

\begin{figure}[t]\centering
\includegraphics[width=1.0\linewidth]{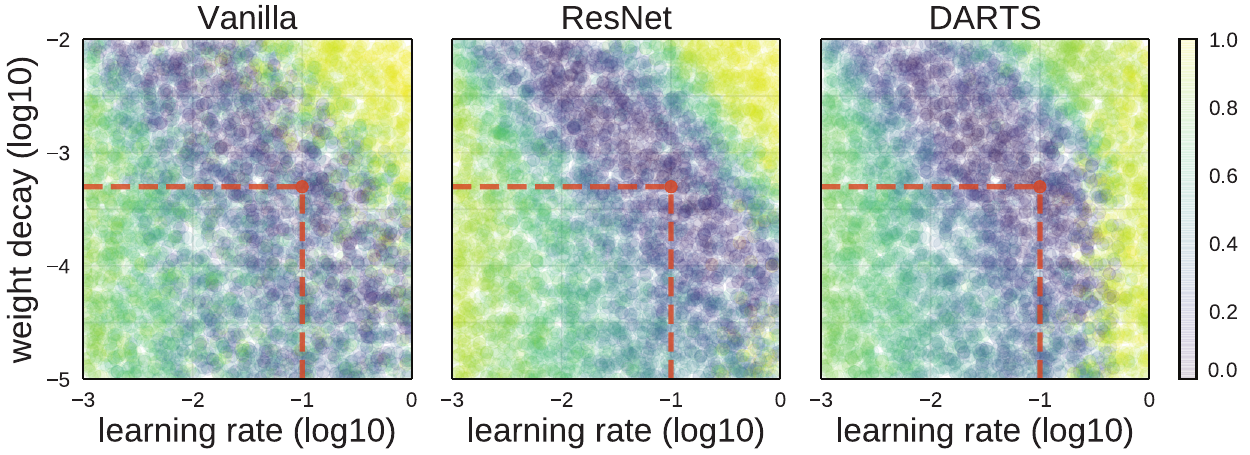}
\caption{\textbf{Learning rate and weight decay.} For each design space, we plot the relative rank (denoted by color) of a randomly sampled model at the ($x$,$y$) coordinate given by the ($lr$,$wd$) used for training. The results are remarkably consistent across the three design spaces (best regions in purple), allowing us to set a single default (denoted by orange lines) for all remaining experiments.}
\label{fig:sanity_lr_wd}\vspace{-1mm}
\end{figure}

\begin{figure}[t]\centering
\includegraphics[width=1.0\linewidth]{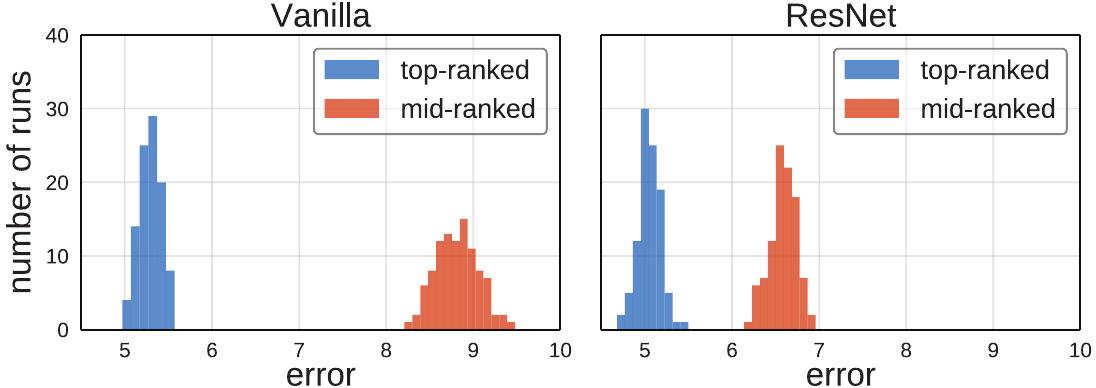}
\caption{\textbf{Model consistency.} Model errors are consistent across 100 reruns of two selected models for each design space, and there is a clear gap between the top and mid-ranked model for each.}
\label{fig:sanity_model_cons}\vspace{-4mm}
\end{figure}

\paragraph{Training schedule.} We use a \emph{half-period cosine schedule} which sets the learning rate via $lr_t = lr(1{+}\cos(\pi{\cdot}t/T))/2$, where $lr$ is the initial learning rate, $t$ is the current epoch, and $T$ is the total epochs. The advantage of this schedule is it has just two hyperparameters: $lr$ and $T$. To determine $lr$ and also weight decay $wd$, we ran a large-scale study with three representative design spaces: \dsname{Vanilla}, \dsname{ResNet}, and \dsname{DARTS}. We train 5k models sampled from each design space for $T{=}$100 epochs with random $lr$ and $wd$ sampled from a log uniform distribution and plot the results in Figure~\ref{fig:sanity_lr_wd}. The results across the three different design spaces are remarkably consistent; in particular, \emph{we found a single $lr$ and $wd$ to be effective across all design spaces}. For all remaining experiments, we use $lr{=}$0.1 and $wd{=}$5e-4.

\paragraph{Training settings.} For all experiments, we use SGD with momentum of 0.9 and mini-batch size of 128. By default we train using $T{=}$100 epochs. We adopt weight initialization from~\cite{He2016} and use standard CIFAR data augmentations~\cite{Lee2015}. For ResNets, our settings improve upon the original settings used in~\cite{He2016}, see Table~\ref{tab:arch}. We note that recent NAS papers use much longer schedules and stronger regularization, \eg see DARTS~\cite{Liu2019}. Using our settings but with $T{=}$600 and comparable extra regularization, we can achieve similar errors, see Table~\ref{tab:best_models}. Thus, we hope that our simple setup can provide a recipe for strong baselines for future work.

\begin{figure}[t]\centering
\includegraphics[width=1.0\linewidth]{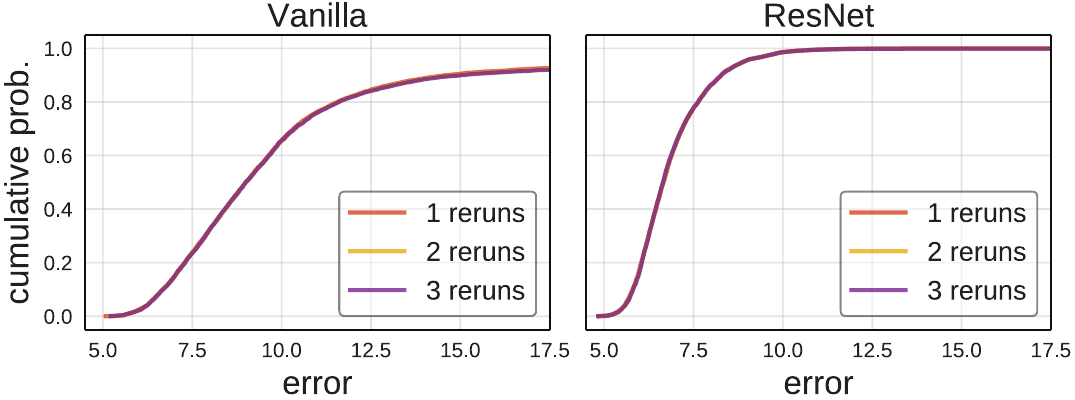}
\caption{\textbf{Trend consistency}. Qualitatively, the trends are consistent across design spaces and the number of reruns. Quantitatively, the $\KS$ test gives $D<0.009$ for both \dsname{Vanilla} and \dsname{ResNet}.}
\label{fig:sanity_trend_cons}
\end{figure}

\paragraph{Model consistency.} Training is a stochastic process and hence error estimates vary across multiple training runs.\footnote{The main source of noise in model error estimates is due to the random number generator seed that determines the initial weights and data ordering. However, surprisingly, even fixing the seed does not reduce the overall variance much due to nondeterminism of floating point operations.} Figure~\ref{fig:sanity_model_cons} shows error distributions of a top and mid-ranked model from the \dsname{ResNet} design space over 100 training runs (other models show similar results). We note that the gap between models is large relative to the error variance. Nevertheless, we check to see if the reliability of error estimates impacts the overall trends observed in our main results. In Figure~\ref{fig:sanity_trend_cons}, we show error EDFs where the error for each of 5k models was computed by averaging over 1 to 3 runs. The EDFs are indistinguishable. Given these results, we use error estimates from a single run in all other experiments.

\begin{figure}[t]\centering
\includegraphics[width=1.0\linewidth]{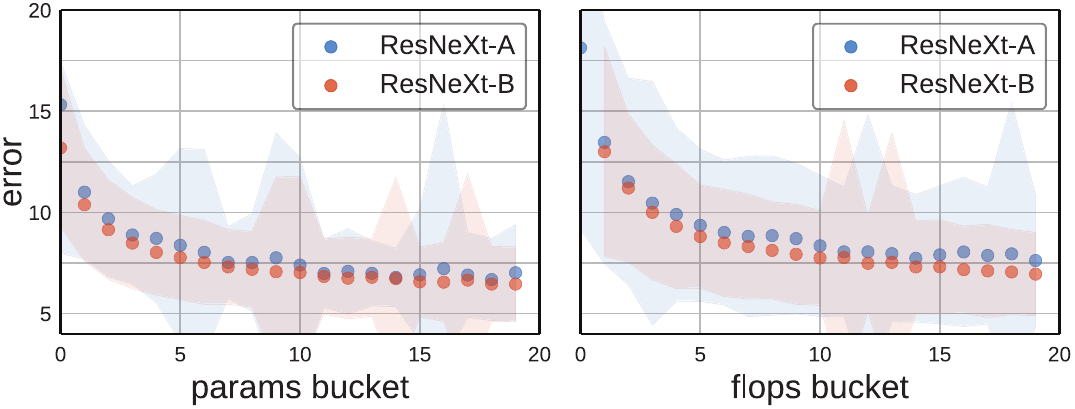}
\caption{\textbf{Bucket comparison.} We split models by complexity into buckets. For each bucket, we obtain a distribution of models and show mean error and two standard deviations (fill). This analysis can be seen as a first step to computing a normalized EDF. }
\label{fig:complexity_bucket_stats}
\end{figure}

\paragraph{Bucket comparison.} Another way to account for complexity is stratified analysis~\cite{Mantel1959}. As in \S\ref{sec:methodology:complexity} we can bin the complexity range into $k$ bins, but instead of reweighing models per bin to generate normalized EDFs, we instead perform analysis within each bin independently. We show the results of this approach applied to our example from \S\ref{sec:methodology:complexity} in Figure~\ref{fig:complexity_bucket_stats}. We observe similar trends as in Figure~\ref{fig:complexity_edfs}. Indeed, bucket analysis can be seen as a first step to computing the normalized EDF (Eqn.~\ref{eq:nedf}), where the data across all bins is combined into a single distribution estimate, which has the advantage that substantially fewer samples are necessary. We thus rely on normalized EDFs for all comparisons.

\begin{figure*}[t]\centering
\begin{minipage}{.49\linewidth}\centering
 \subfloat[\textbf{NAS distribution comparisons} (compare to Fig.~\ref{fig:nas_edfs}).
 \label{fig:in_nas_edfs}]{\includegraphics[width=1\linewidth]{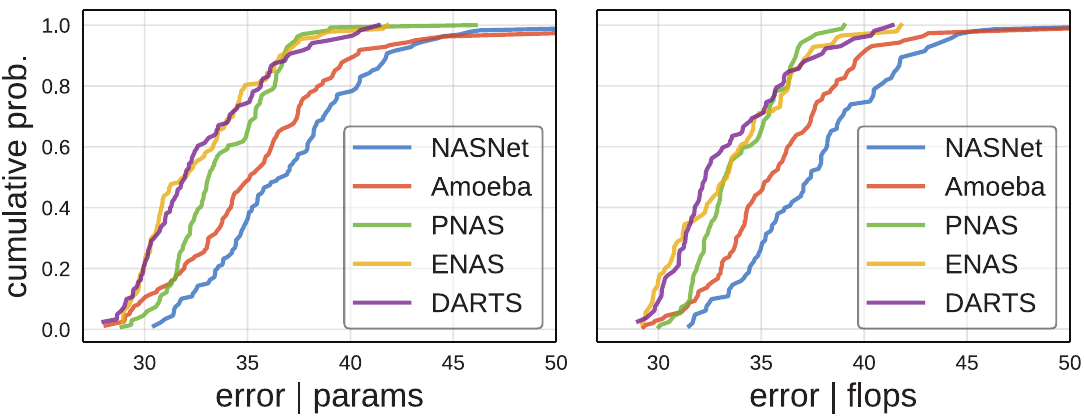}}\\
 \subfloat[\textbf{NAS random search efficiency} (compare to Fig.~\ref{fig:nas_rs}).
 \label{fig:in_nas_rs}]{\includegraphics[width=1\linewidth]{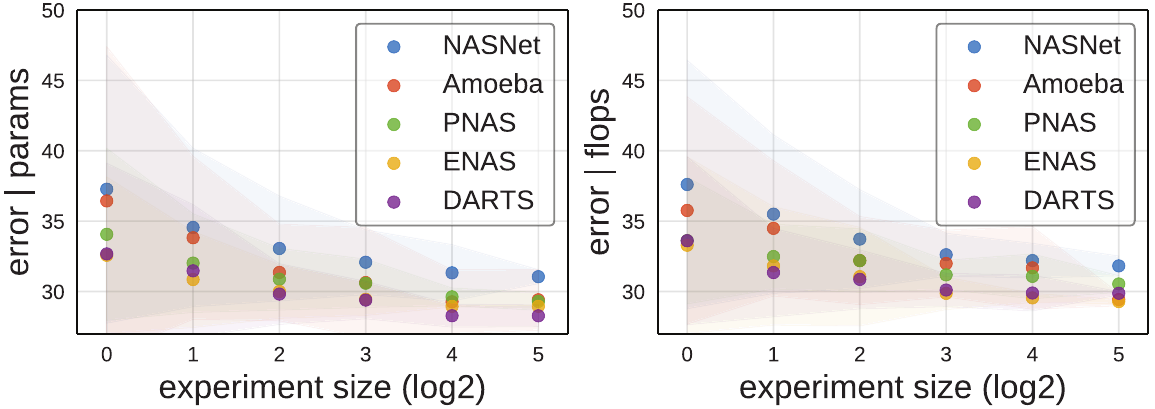}}
\end{minipage}\hspace{2mm}
\begin{minipage}{.49\linewidth}\centering
 \subfloat[\textbf{NAS \vs standard design spaces} (compare to Fig.~\ref{fig:nas_std_edfs}).
 \label{fig:in_nas_std_edfs}]{\includegraphics[width=1\linewidth]{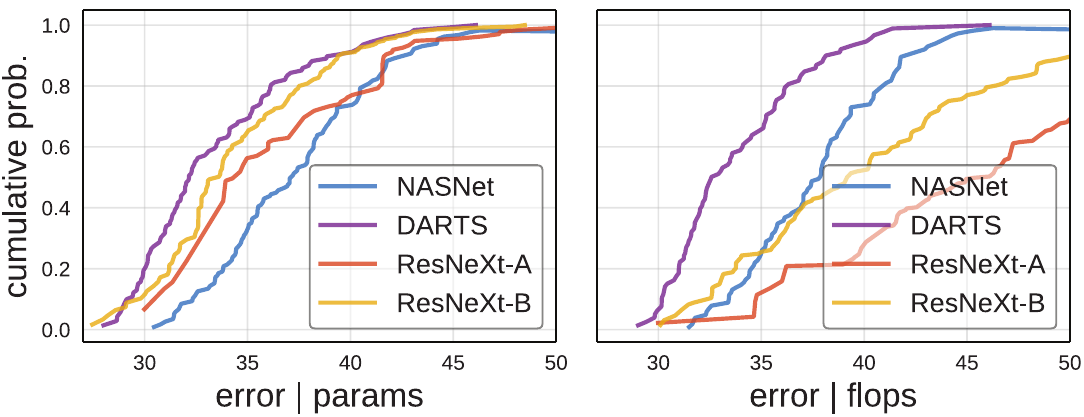}}\\
 \subfloat[\textbf{Learning rate and weight decay} (compare to Fig.~\ref{fig:sanity_lr_wd}).
 \label{fig:in_lr_wd}]{\includegraphics[width=1\linewidth]{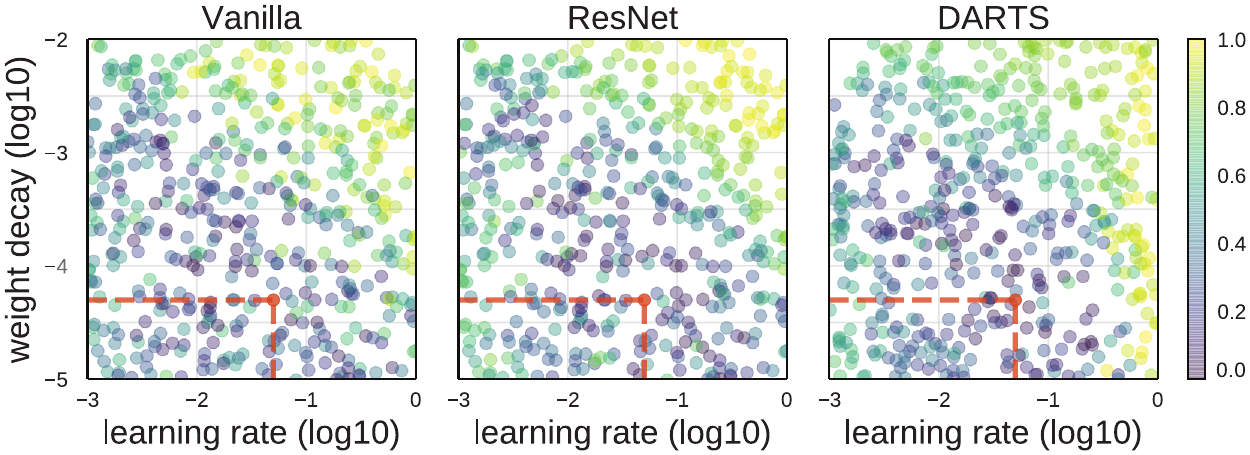}}
\end{minipage}\\[1mm]
\caption{\textbf{ImageNet experiments}. In general, the results on ImageNet closely follow the results on CIFAR.}
\label{fig:in_results}\vspace{-2mm}
\end{figure*}

\section*{Appendix B: ImageNet Experiments}

We now evaluate our methodology in the large-scale regime of ImageNet~\cite{Deng2009} (IN for short). In particular, we repeat the NAS case study from \S\ref{sec:nas} on IN. We note that these experiments were performed after our methodology was finalized, and we ran these experiments exactly once. Thus, IN can be considered as a test case for our methodology.

\paragraph{Design spaces.} We provide details about the IN design spaces used in our study. We use the same \textit{model families} as in \S\ref{sec:ds:instantiation} and \S\ref{sec:nas:ds}. The precise \textit{design spaces} on IN are as close as possible to their CIFAR counterparts. We make only two necessary modifications: adopt the IN stem from DARTS~\cite{Liu2019} and adjust NAS width and depth values to $w\in\{32,48,64,80,96\}$ and $d\in\{6,10,14,18,22\}$. We keep the allowable hyperparameter values for ResNeXt design spaces unchanged from Table~\ref{tab:design_spaces} (our IN models have 3 stages). We further upper-bound models to 6M parameters or 0.6B flops which gives models in in the mobile regime commonly used in the NAS literature. The \textit{model distributions} follow \S\ref{sec:ds:instantiation} and \S\ref{sec:nas:ds}. For \textit{data generation}, we train $\app$100 models on IN for each of the five NAS design spaces in Table~\ref{tab:nas_details} and two ResNeXt design spaces in Table~\ref{tab:design_spaces}. Note that to stress test our methodology we choose to use the minimal number of samples (see \S\ref{sec:methodology:minimal} for discussion).

\paragraph{NAS distribution comparisons.} In Figure~\ref{fig:in_nas_edfs} we show normalized EDFs for the NAS design spaces on IN. We observe that the general EDF shapes match their CIFAR counterparts in Figure~\ref{fig:nas_edfs}. Moreover, the relative ordering of the NAS design spaces is consistent between the two datasets as well. This provides evidence that current NAS design spaces, developed on CIFAR, are transferable to IN.

\paragraph{NAS random search efficiency.} Analogously to Figure~\ref{fig:nas_rs}, we simulate random search in the NAS design spaces on IN and show the results in Figure~\ref{fig:in_nas_rs}. Our main findings are consistent with CIFAR: (1) random search efficiency ordering is consistent with EDF ordering and (2) differences in design spaces alone result in differences in performance.

\paragraph{Comparison to standard design spaces.} We next follow the setup from Figure~\ref{fig:nas_std_edfs} and compare NAS design spaces to standard design spaces on IN in Figure~\ref{fig:in_nas_std_edfs}. The main observation is again consistent: standard design spaces \emph{can be} comparable to the NAS ones. In particular, \dsname{ResNeXt-B} is similar to \dsname{DARTS} when normalizing by params (left), while NAS design spaces outperform the standard ones by a considerable margin when normalizing by flops (right).

\paragraph{Discussion.} Overall, our IN results closely follow their CIFAR counterparts. As before, our core observation is that the \emph{design} of the design space can play a key role in determining the potential effectiveness of architecture search. These experiments also demonstrate that using 100 models per design space is sufficient to apply our methodology and strengthen the case for its feasibility in practice. We hope these results can further encourage the use of distribution estimates as a guiding tool in model development.

\paragraph{Training settings.} We conclude by listing detailed IN training settings, which follow the ones from Appendix A unless specified next. We train our models for 50 epochs. To determine $lr$ and $wd$ we follow the same procedure as for CIFAR (results in Figure~\ref{fig:in_lr_wd}) and set $lr{=}$0.05 and $wd{=}$5e-5. We adopt standard IN data augmentations: aspect ratio~\cite{Szegedy2015}, flipping, PCA~\cite{Krizhevsky2012}, and per-channel mean and SD normalization. At test time, we rescale images to 256 (shorter side) and evaluate the model on the center 224$\x$224 crop.

\section*{Acknowledgements}
{\small We would like to thank Ross Girshick, Kaiming He, and Agrim Gupta for valuable discussions and feedback.}

{\small\bibliographystyle{ieee}\bibliography{nms}}

\end{document}